\documentclass[10pt,twocolumn,letterpaper]{article}
\usepackage[pagenumbers]{cvpr} 
\usepackage[T1]{fontenc}
\usepackage{amsmath,amssymb,booktabs,multirow,colortbl}
\definecolor{oursrow}{RGB}{232,240,251}
\makeatletter\newcommand{\oursrow}{\makebox[0pt][l]{\color{oursrow}\rule[-\dp\@arstrutbox]{\linewidth}{\dimexpr\ht\@arstrutbox+\dp\@arstrutbox\relax}}}\makeatother
\usepackage{microtype}
\usepackage{cuted,needspace}
\definecolor{cvprblue}{rgb}{0.21,0.49,0.74}
\usepackage[pagebackref,breaklinks,colorlinks,allcolors=cvprblue]{hyperref}
\def\confName{CVPR}
\def\confYear{2027}
\title{VideoRSI: Verifiable Self-Improvement for Code-to-Video Generation}
\author{%
Cheng Chen$^{1,2}$\quad
Yi Tan$^{1}$\quad
Tianrun Chen$^{4,5}$\quad
Qi Guo$^{6}$\quad
Xinyao Liao$^{1}$\quad
Deyi Ji$^{5,7}$\\
Qi Zhu$^{5,7}$\quad
Lanyun Zhu$^{8}$\quad
Keyu Wen$^{9}$\quad
Zhenshan Tan$^{3}$\quad
Guosheng Lin$^{1}$\quad
Fayao Liu$^{2}$\\[2pt]
{\footnotesize $^{1}$Nanyang Technological University\quad
$^{2}$Institute of Advanced Intelligence and Computing, A*STAR}\\
{\footnotesize $^{3}$Nanjing University of Information Science and Technology\quad
$^{4}$Zhejiang University\quad
$^{5}$Moxin Technology}\\
{\footnotesize $^{6}$Inner Mongolia University\quad
$^{7}$University of Science and Technology of China\quad
$^{8}$Tongji University\quad
$^{9}$Fudan University}
}
\begin{document}
\maketitle
\begin{strip}
\centering
\includegraphics[width=\textwidth]{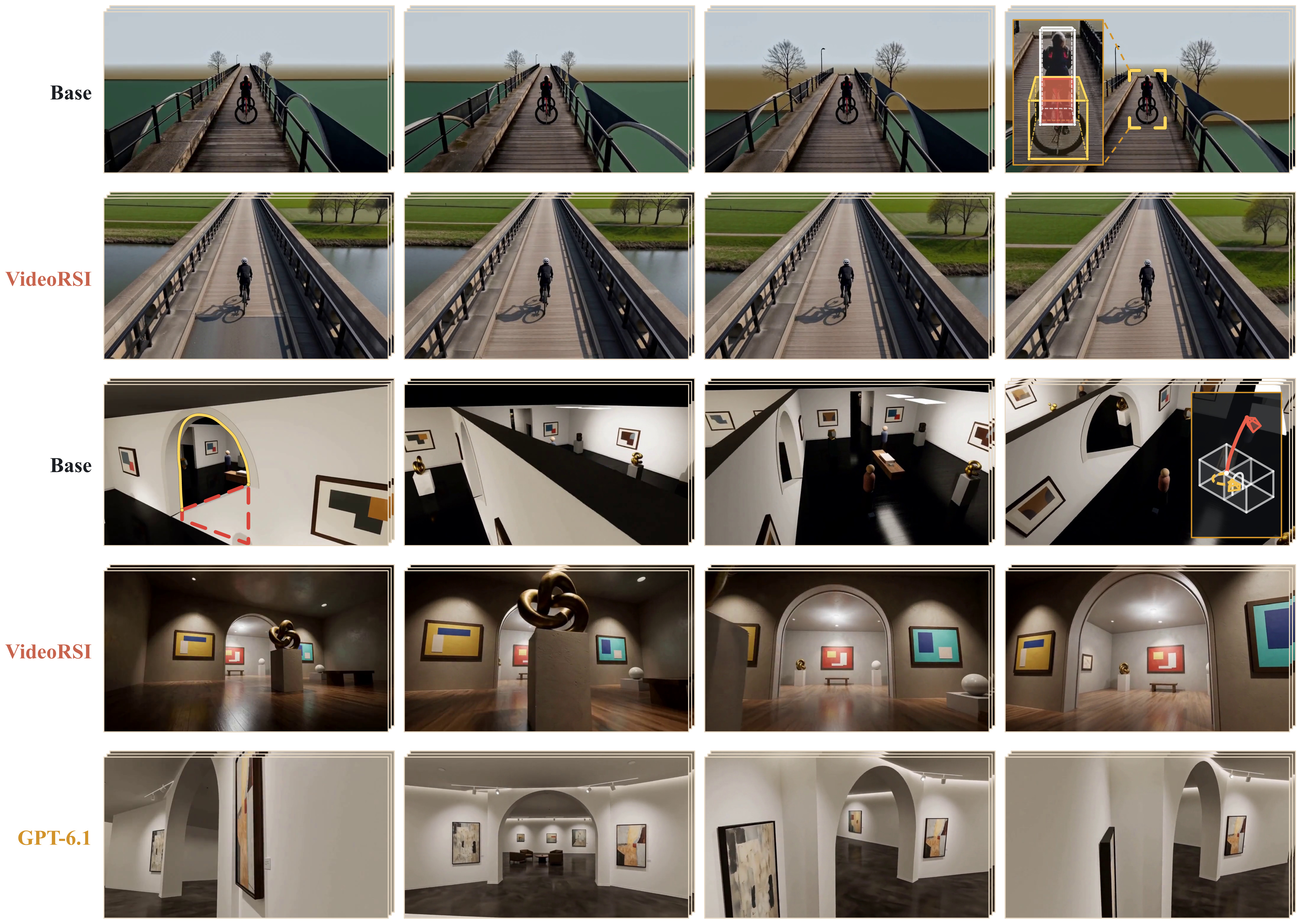}
\vspace{-14pt}
\captionsetup{hypcap=false}\captionof{figure}{Final videos generated from program proxies, four frames per row. After self-improvement, VideoRSI builds more plausible scenes. Rows 1--2: the Base director (Qwen3.8-27B) shows the classic bicycle-wheel clipping, which VideoRSI avoids. Rows 3--5: Base builds wrong walls and its camera fails to sweep around the gallery and then look into the rear room; VideoRSI follows this move, and more closely than GPT-6.1 Sol (Extra-High Reasoning).}\label{fig:1}
\vspace{4pt}
\end{strip}

\captionsetup{hypcap=true}
\begin{abstract}
Executable environments let coding agents improve themselves; the key is reliable program verification. In code-to-video generation, an LLM director writes a 3D scene program that a video model renders realistically. Coupling geometry, physics, motion and camera, this complex visual task is hard to verify. To make it verifiable for self-improvement, we ask what to verify and how. We observe that the generator treats the proxy as a hard 4D constraint and the prompt as a soft appearance prior: program-level 4D errors pass irreversibly into the video, yet stay invisible to VLM judges on 2D frames. We therefore verify at the program level: VideoRSI executes each program and maintains a unified 4D state of objects and camera, on which all spatiotemporal constraints are checked. As a non-compensatory filter, the verifier drives self-improvement at two levels: rejection-sampling fine-tuning on verified self-generated trajectories updates the weights, and verifier-gated harness evolution refines the workflow. The resulting \mbox{Qwen3.8-27B} director yields much better videos with equally rich scenes, approaching frontier closed-source models (\mbox{GPT-6.1}). Project page: \url{https://steller-chen.github.io/VideoRSI/}.
\end{abstract}

\section{Introduction}

Executable environments offer a promising route to self-improvement for coding agents: a model writes programs, observes their execution, localizes errors and learns from successful attempts \cite{star,rest,swegym}. This route rests on one premise: the environment must reliably verify whether a program has truly improved. For complex visual programming, this is hard. Program quality depends not only on appearance but also on interrelated properties such as physical plausibility, spatial relations, dynamic consistency and camera control, and how to verify them reliably remains understudied.

We study this problem in code-to-video generation. Recent works \cite{cvm,videococo} use a language model as a ``director'' that writes an executable 3D scene program, and a video generation model then turns the program-rendered proxy video into a realistic video. This makes layout, motion, interaction and camera control programmable, combining the precise control of graphics with the realism of generative video. This young direction still relies on commercial closed-source models as directors. For open-source directors to improve, we ask: \textbf{\textit{what should be verified, and how, to reliably guide the self-improvement of a code-to-video director?}}

\begin{figure}[t]
\centering
\includegraphics[width=\columnwidth]{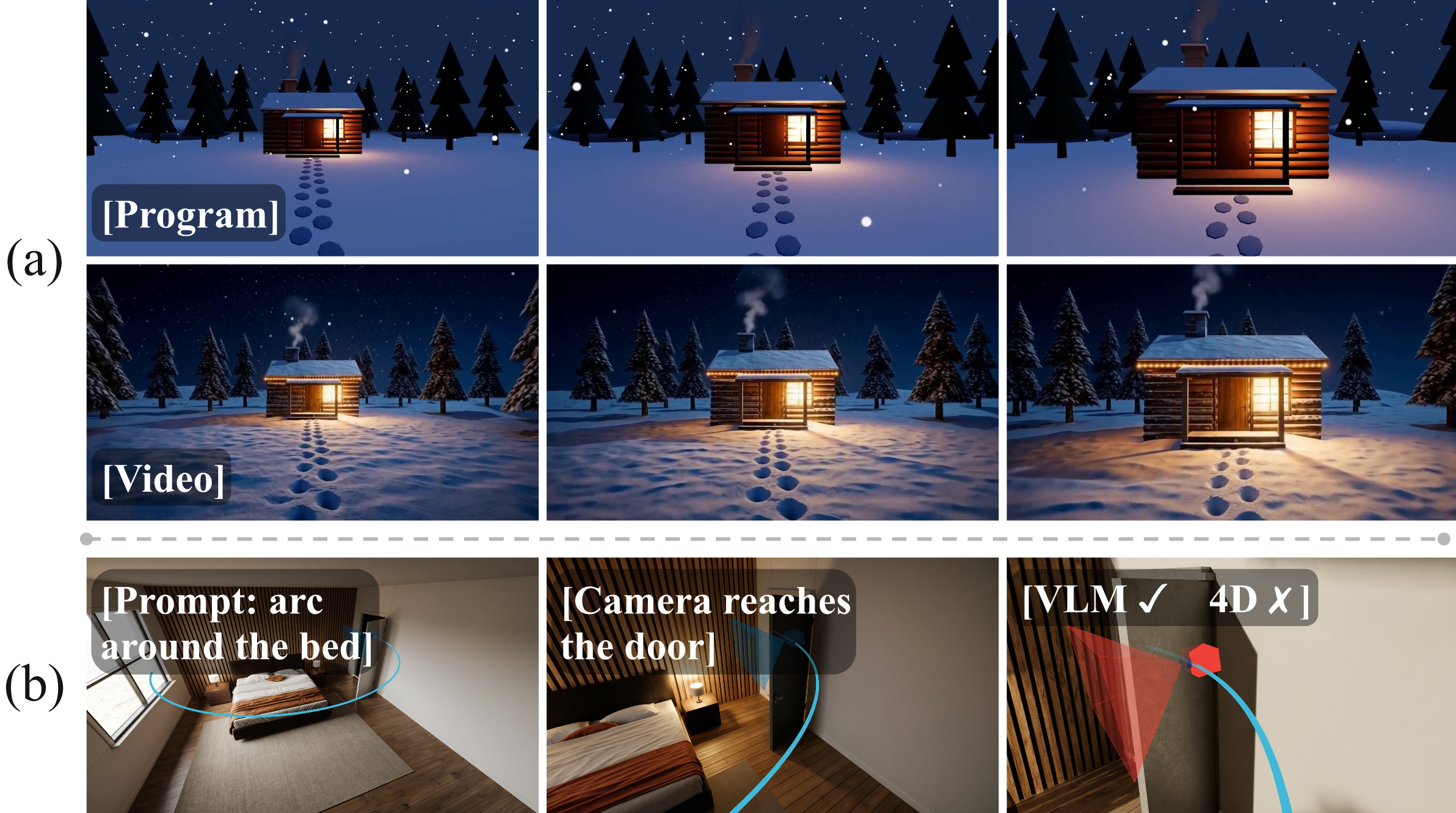}
\caption{(a) Top: program render. Bottom: generated video. Structure and camera motion carry over, while the video model rewrites appearance. (b) The prompt asks the camera to circle the bed, but the program camera passes through the door. A VLM viewing the frames finds no problem; our 4D check catches the collision.}\label{fig:2}
\end{figure}

\emph{What} to verify depends on what the director actually controls in the final video. The video generator is conditioned on both the proxy video and the text prompt, and we observe that it treats the two very differently. Content is negotiable: appearance is rewritten freely, and details described only in the prompt appear in the video even when the program omits them. 4D structure is not: spatial layout, motion trajectories, event timing and camera motion follow the proxy even when the prompt asks for something else, and 4D errors, such as a camera passing through a door, survive into the video (Fig.~\ref{fig:2}). The generator thus re-renders content but executes 4D structure as given, so the final video can be no better in 4D than the program that drives it. What to verify is therefore 4D quality, while keeping the scene rich.

\looseness=-1 As for \emph{how}, we first study the common VLM-as-a-Judge approach \cite{mllmjudge,videoscore}. We find that clear 4D errors, such as the camera passing through the door in Fig.~\ref{fig:2}b, go unnoticed when it scores keyframes (details in the appendix). A few frames do not determine 3D contact, motion continuity or camera geometry, so the judge can only guess them \cite{whatsup}. 4D quality should therefore be verified on the program itself, instead of guessed from a few frames.

Answering both questions, we propose VideoRSI, a framework that improves open-source coding agents through executable 4D verification. We run the program in a headless browser and maintain a 4D state that tracks every object and the camera over time. On this state, we can easily check geometric plausibility, camera control, motion readability and spatial grounding. The verifier returns per-dimension scores to select better programs and diagnostics that locate each error to guide its fix. The checks are efficient and reproducible, and each dimension is gated on its own, so that, for example, good framing cannot hide a collision.

\looseness=-1 With this verifier, VideoRSI improves the director in two stages: first its weights, then its harness. Starting from Qwen3.8-27B \cite{qwen}, we filter the model's own trajectories with the 4D constraints, dimension by dimension, and apply rejection-sampling fine-tuning (RS-SFT) \cite{rft,rest} so that the model learns from its better experience. In Harness Evolution, the updated model proposes edits to its instructions, skills and execution strategies, and an edit is kept only if it improves 4D quality over the current one on the same prompts.

The two stages are complementary: RS-SFT improves the scene-programming ability of the model, and Harness Evolution further improves scene quality. Because both are driven by the same 4D signal, one verifier improves both what the model learns and how the agent works, strengthening its spatial, motion and camera control.

\looseness=-1 We evaluate VideoRSI on the final videos across task categories and a public benchmark \cite{codebench}. The self-improved director substantially raises the 4D quality of its scenes over the base model and generalizes across task categories; against GPT-6.1 (Extra-High Reasoning), the 27B model achieves competitive scene and camera control (Fig.~\ref{fig:1}).

Our contributions are as follows:
\begin{itemize}\setlength{\itemsep}{2pt}\setlength{\parskip}{0pt}\setlength{\topsep}{2pt}
\item We reveal an asymmetry in code-to-video generation: the video generator freely re-renders content but faithfully executes 4D structure. It fills in details the program omits, yet keeps the program's layout, motion, timing and camera, errors included, so the video is no better in 4D than the program, a limit frame-based VLM rewards cannot see.
\item We introduce executable 4D verification, which checks the dynamic scene itself rather than its frames: object-level 4D states read during execution yield efficient, reproducible per-dimension scores and localized diagnostics.
\item We use the verifier as a gate at two levels: per-dimension constraints select self-generated trajectories for RS-SFT, and harness edits must raise 4D quality without hurting geometry, so simplifying a scene does not pay off.
\end{itemize}

\section{Related Work}

\begin{figure*}[t]
\centering
\includegraphics[width=\textwidth]{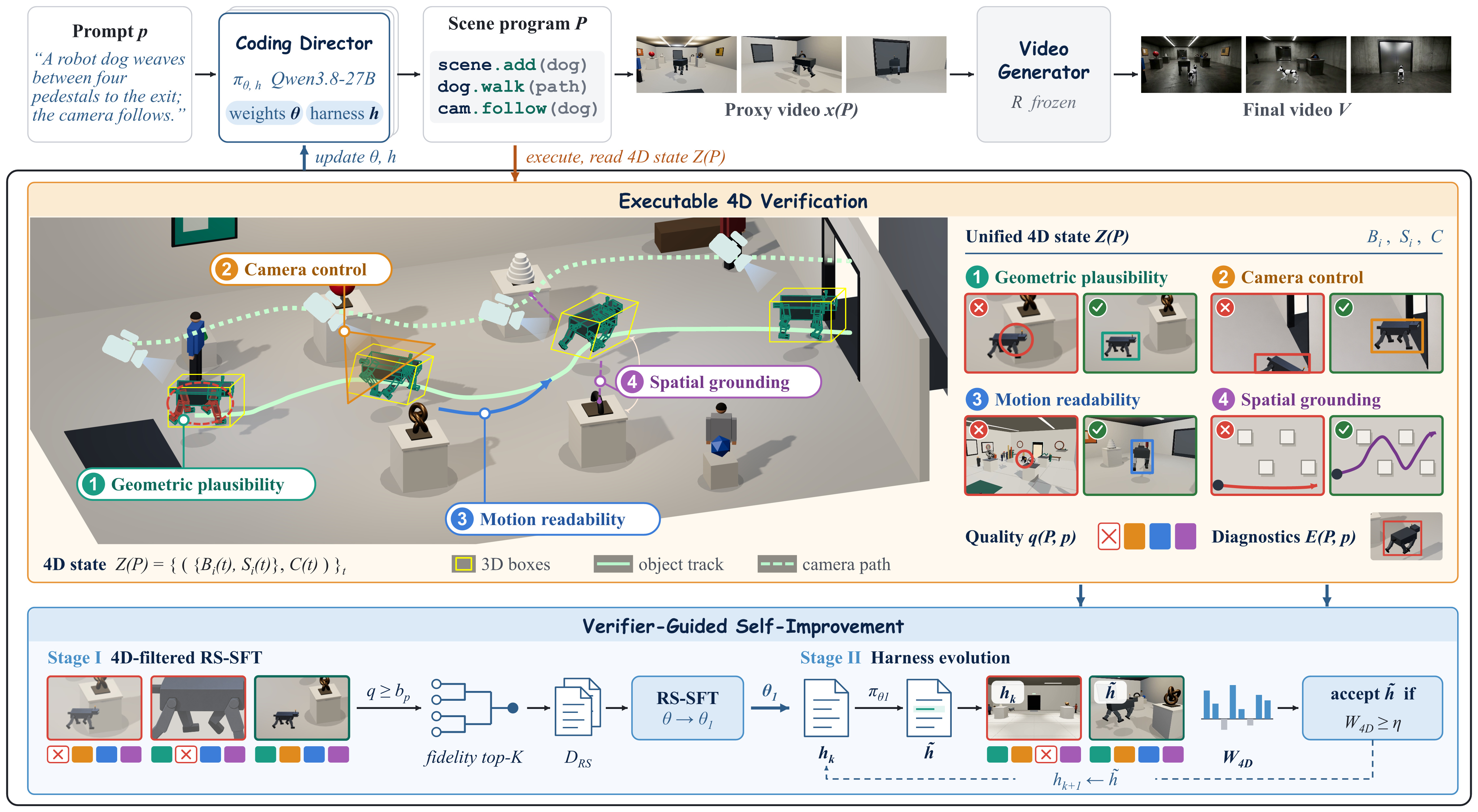}
\caption{VideoRSI overview. Execution yields a unified 4D state of the scene; four checks return per-dimension scores and diagnostics. Constraint-aware selection gates every dimension before ranking task fidelity, so simplifying the scene cannot raise the score. Stage I fine-tunes the weights on verified trajectories (what the director learns); Stage II evolves its harness under the same verifier (how it works).}\label{fig:3}
\end{figure*}

\textbf{Controllable video generation through programs.} CameraCtrl conditions generation on camera trajectories \cite{camera}, and VACE unifies reference, editing and masking controls \cite{vace}. VideoDirectorGPT plans layouts with an LLM \cite{videodirectorgpt}, GPT4Motion turns language plans into Blender physics scripts \cite{gpt4motion}, SceneCraft synthesizes 3D scenes as code \cite{scenecraft}, and BlenderAlchemy edits them with a VLM evaluator \cite{blenderalchemy}. CVM connects executable proxies to generative appearance through early-step conditioning \cite{cvm}, VideoCoCo uses Blender drafts as process-level intermediates \cite{videococo}, and Video-Coder trains video-programming agents through SFT and agentic RL with execution and video rewards \cite{videocoder}. These works study how programs control generation; VideoRSI instead studies how a director improves its own programs with verifiable feedback under a frozen video generator.

\textbf{Self-improvement and verification feedback.} STaR and later self-training methods learn from feedback-filtered self-generated outputs \cite{star,rest}, and SWE-Gym trains software-engineering agents with executable environments and verifiers \cite{swegym}. Other work improves strategies outside the weights: ADAS searches agent programs \cite{adas}, the Darwin G\"odel Machine evaluates self-modifying agents \cite{dgm}, GEPA evolves prompts through trajectory reflection \cite{gepa}, and LIT co-evolves director and meta agents for video production \cite{lit}. All of these routes depend on feedback quality. VBench \cite{vbench}, VideoScore \cite{videoscore}, VideoPhy \cite{videophy} and T2V-CompBench \cite{t2vcompbench} score videos from pixels, yet vision-language models struggle with basic spatial relations \cite{whatsup}, and mismatched rewards invite reward hacking \cite{rewardgaming}. VideoRSI instead reads object-level 4D states from executed programs, using one verifier to filter trajectories and accept harness edits.

\section{Method}

\subsection{Problem Formulation}

Given a prompt $p$, an open-source director $\pi_{\theta,h}(P\mid p)$, parameterized by model weights $\theta$ and harness $h$, generates an executable Three.js \cite{threejs} program $P$. The program constructs a complete dynamic 3D scene, including objects, geometry, appearance, layout, interactions, time-varying motion and camera behavior. Executing $P$ yields a proxy video $x(P)$. A video generator $R$ then renders the final video $V=R\bigl(x(P),I_0(P,p),p\bigr)$ from the proxy, the prompt and a reference image $I_0(P,p)$ edited from the proxy first frame.

VideoRSI (Fig.~\ref{fig:3}) freezes the video generator and improves scene programs so that they better organize the layout, events and camera presentation of final videos. We guide this improvement with verifiable program attributes and test its effects in the generated output.

\subsection{Executable 4D Verification}

\noindent\textbf{Executable 4D Scene Representation.} Dynamic scenes are explicitly defined by programs. We can therefore read scene states directly during execution, rather than infer geometry and motion from images. Executing $P$ in a headless browser, we collect a unified object-level 4D representation:
\begin{equation}
\mathcal Z(P)=\bigl\{\,\bigl(\{(B_i(t),S_i(t))\}_{i\in\mathcal I},\ C(t)\bigr)\,\bigr\}_{t\in\mathcal T}.
\end{equation}
Here $B_i(t)$ is the world-space 3D bounding box of object $i$, $S_i(t)$ its projected screen box, and $C(t)$ the camera pose and projection parameters. Object identities persist over time. These encode world-space geometry, screen-space visibility and composition, and viewpoint changes, respectively. Temporal association makes $\mathcal Z(P)$ a common representation of scene layout, object motion and camera motion.

\looseness=-1 \noindent\textbf{Controllability-Aligned 4D Verification.} Verification checks relations between objects, objects over time, and objects and cameras. All three are directly available from $\mathcal Z(P)$ and determine spatial layout, dynamic events and viewpoints. We measure four complementary dimensions.

Geometric plausibility checks spatial and temporal contradictions, such as interpenetration, abrupt position jumps, or unsupported hovering or rising of passive, gravity-constrained objects. It detects observable state violations rather than certifying complete physical correctness. Camera control checks reasonable camera placement and effective framing: camera penetration, excessive roll, missing subjects, or subjects too small or too large. Motion readability combines screen size, visible duration and motion relative to the camera to assess whether required actions are legible. Spatial grounding maps named objects and relative-position requirements to the scene: a VLM links textual references to program objects, while 3D geometry determines position and direction relations.

Checks apply only when supported by the task semantics and scene conditions, avoiding penalties for intentional actions or camera motion. They are complementary tests of geometry, motion and observation in the same scene, rather than independent visual scorers. Results form a higher-is-better quality vector:
\begin{equation}
\mathbf q(P,p)=\bigl[q_{\rm geo},q_{\rm cam},q_{\rm mot},q_{\rm spa}\bigr]^{\!\top}.
\end{equation}
Checks also return violation records $\mathcal E(P,p)$ with object identity, time and location. Quantitative measures evaluate candidates; diagnostics explain specific problems and guide subsequent strategy changes.

\noindent\textbf{Constraint-Aware Quality Selection.} A weighted scalar can hide critical weaknesses: good framing cannot offset a motion anomaly, and removing actions should not count as improving dynamics. We therefore apply per-dimension quality gates before comparing task fidelity. For candidate programs $\mathcal P_p$ for the same prompt, define:
\begin{equation}
\mathcal F_p=\{P\in\mathcal P_p:\ \mathbf q(P,p)\ge\mathbf b_p\},\label{eq:feasible}
\end{equation}
The per-dimension standards $\mathbf b_p$ are determined from better candidates for the same prompt with tolerances; $\ge$ applies per dimension, so every dimension must pass. Pairwise comparisons then select feasible programs more faithful to the requested subjects, events and scene intent. The 4D constraints reject evident defects; fidelity comparison distinguishes actual task completion. Together they discourage score gains from deleting content or simplifying motion.

\looseness=-1 This division limits the visual judge: it compares task semantics and content completion, while executable states determine spatial, motion and camera geometry. No program metric is assumed to represent final-video quality; output evaluation tests the selection principle independently.

\subsection{Verifier-Guided Self-Improvement}

\looseness=-1 The same signal supports two operations. Weight learning filters high-quality self-generated trajectories using the quality constraints. Harness optimization compares and accepts execution strategies using 4D feedback. We update weights first, then evolve the harness with those weights fixed.

\noindent\textbf{Stage I: 4D-Filtered Rejection-Sampling Fine-Tuning.} For each training prompt $p$, the base director generates candidate programs $\mathcal P_p$ and agent trajectories $\tau(P)$. The per-dimension constraints in Eq.~(\ref{eq:feasible}) reject clear spatiotemporal weaknesses. Pairwise task-fidelity comparisons rank the feasible trajectories:
\begin{equation}
\mathcal D_{\mathrm{RS}}=
\bigcup_{p\in\mathcal Q_{\mathrm{train}}}
\operatorname{TopK}_{\mathrm{fid}}
\bigl(\{\tau(P):P\in\mathcal F_p\}\bigr).
\end{equation}
Here $\operatorname{TopK}_{\mathrm{fid}}$ selects the top $K$ trajectories through a prompt-fidelity tournament. We split selected trajectories into training windows $(c,y)$ and apply \mbox{RS-SFT}:
\begin{equation}
\theta_1=\arg\min_{\theta}\;\mathbb E_{(c,y)\sim\mathcal W(\mathcal D_{\mathrm{RS}})}
[-\log\pi_{\theta}(y\mid c)].
\end{equation}
\looseness=-1 $\mathcal W$ maps trajectories to windows of execution context $c$ and supervised output $y$. The 4D constraints remove spatiotemporal defects; task fidelity selects programs that fulfill the scene request. The model learns from its valid experiences rather than optimizing a single verifier score.

\noindent\textbf{Stage II: Feedback-Guided Harness Evolution.} With $\theta_1$ fixed, we search harness instructions, skills and execution strategies. Given $h_k$, the updated model reads historical trajectories, 4D evaluations and diagnostics to propose executable harness edits. We evaluate each candidate against its parent on the same development prompts and compute budget. Per-task normalized 4D quality changes yield a net win rate; staged evaluation reduces selection noise.

A candidate must meet the improvement threshold while preserving geometric plausibility and delivery stability. Formally, $\tilde h$ must satisfy:
{\small
\begin{equation}
W_{4D}(\tilde h,h_k)\ge\eta,\;\; \bar q_{\rm geo}(\tilde h)\ge\bar q_{\rm geo}(h_k),\;\; \bar d(\tilde h)\ge\bar d(h_k).
\end{equation}}
\looseness=-1 $W_{4D}$ is the candidate's net 4D win rate against its parent on development prompts; $\bar q_{\rm geo}$ and $\bar d$ denote mean geometry and delivery, and $\eta$ is the admission threshold. Passing candidates update the harness; otherwise the parent remains. Search thus optimizes consistent scene improvement on matched tasks, rather than a favorable isolated generation.

These stages sequentially improve model parameters and external execution strategies through program-level feedback. Held-out tasks and automatic and human final-video evaluation test whether the gains reach the output.

\section{Experiments}

\begin{figure*}[t]
\centering
\includegraphics[width=0.9\textwidth]{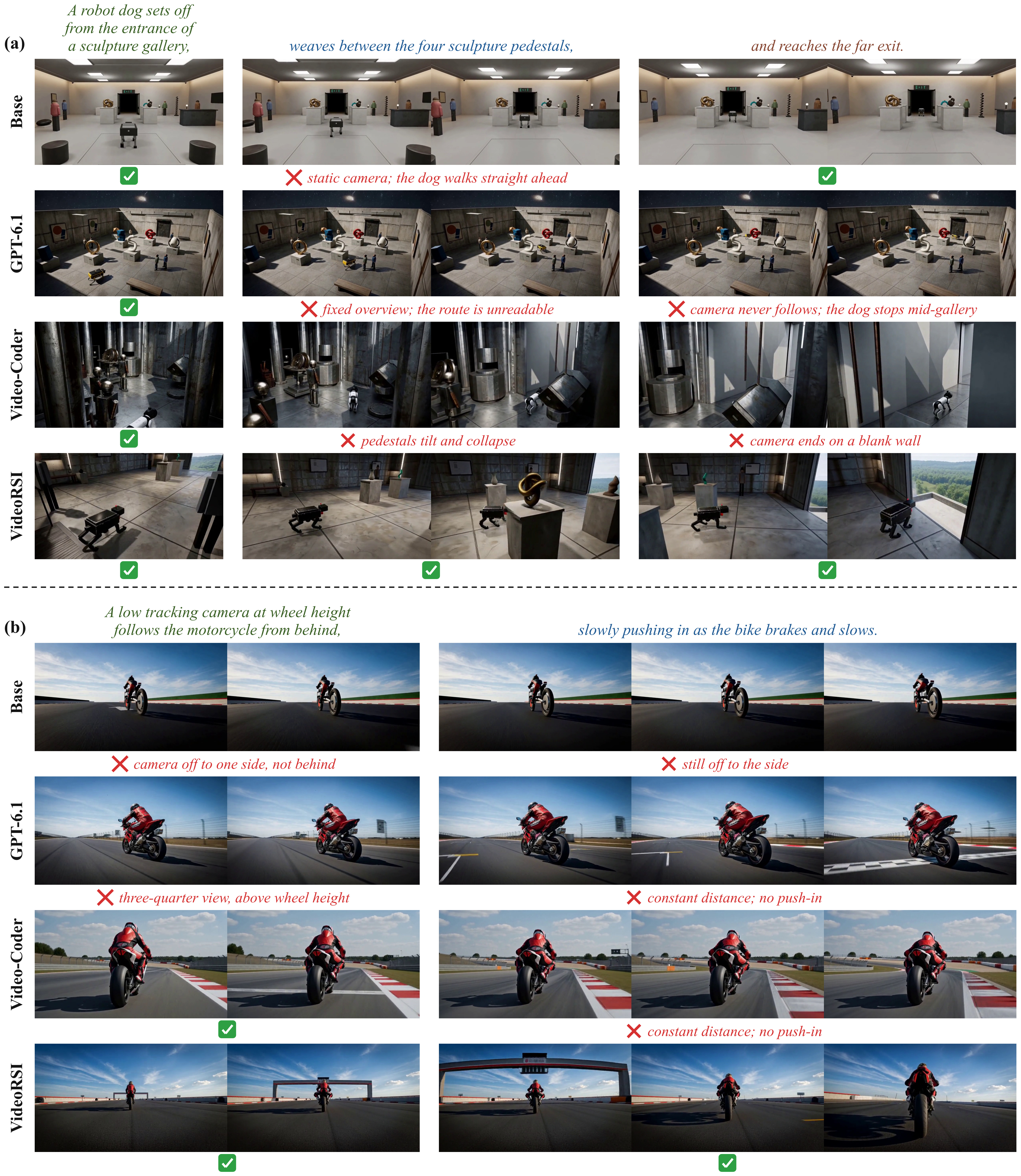}
\caption{Final-video comparisons: (a) a dog weaves between pedestals and reaches the exit; (b) low rear motorcycle tracking followed by a push-in. Colors separate instruction stages; ticks and crosses annotate execution.}\label{fig:4}
\end{figure*}

\subsection{Implementation Details}

\noindent\textbf{Training setup.} We build the director on Qwen3.8-27B \cite{qwen}. Following CVM \cite{cvm}, our task collection covers 10 categories of complex code-to-video tasks; 60 tasks from two entirely unseen categories test cross-category generalization. Sampling the base director on the training split yields 804 self-generated agent trajectories. The collection scale is comparable to agent-training studies such as SWE-Gym \cite{swegym}, but interactions are longer and multimodal, averaging about 45 turns of coding, execution and visual observation. Training uses LoRA \cite{lora} rank 64, a learning rate of $2\times10^{-5}$ and 32 NVIDIA A800 GPUs. Qwen-Image-Edit-2511 \cite{qwenedit} produces reference images; frozen MiniMax-H3 \cite{minimax} synthesizes final videos.

\noindent\textbf{Long-context training and KV-cache implementation.} A single agent trajectory reaches about 220K tokens of interleaved code, tool outputs and rendered frames, far beyond a trainable context. We cut each trajectory into turn-aligned windows of at most 16K tokens that never split a tool call from its response, and apply an assistant-only loss to the final target turn. Logits are computed only at supervised positions, which avoids the full-vocabulary projection over the history and cuts activation memory. Training uses ZeRO-3 sharding \cite{zero} and gradient checkpointing \cite{checkpoint}. Text-only and vision-language samples share one forward path, so every rank gathers sharded parameters in the same order; coding and visual turns can then share a batch without mismatched collectives. At inference, agent histories are append-only: session-affinity routing keeps each episode on one vLLM \cite{vllm} replica, and prefix KV-cache reuse \cite{prefixcache} skips recomputing the shared prefix at every turn. Load-matched evaluation compares harnesses under the same serving load, so cache contention does not bias the comparison.

\subsection{Comparison with State-of-the-Art Methods}

\textbf{Settings.} We compare three representative director types to assess scene-programming control and final-video quality.

\textbf{Base director.} Qwen3.8-27B with the original CVM \cite{cvm} harness uses the same scene and video pipeline, without additional weight training or harness optimization, isolating the benefit of verifiable self-improvement.

\looseness=-1 \textbf{Recent code-to-video agent.} Video-Coder \cite{videocoder} trains a Blender-based coding agent with SFT and agentic RL. We run its public model with the same prompts and MiniMax-H3 setup to compare training paradigms.

\textbf{Frontier commercial director.} GPT-6.1 \cite{openai} (Extra-High Reasoning) uses the same harness and generation pipeline as Base to measure the gap to a frontier director.

\looseness=-1 Our 82-prompt main benchmark combines 60 unseen-category tasks with all 22 tasks of the public CodeVideoBench \cite{codebench}, covering complex camera control, dynamic interactions and 3D scene generation. The development prompts used in Harness Evolution are disjoint from all prompts in this benchmark.

\looseness=-1 \noindent\textbf{Evaluation metrics.} To specifically assess camera control, we compute a VGGT-based \cite{vggt} Camera Instruction Accuracy (CIA): each camera instruction is decomposed into checklist items, judged from estimated camera poses and from subject positions given by Grounding DINO \cite{groundingdino} and SAM~2 \cite{sam2}, and CIA is the fraction satisfied. Furthermore, to evaluate whether the final video meets the remaining requirements, we propose two metrics. Scene Specification Fidelity (SSF) assesses scene realization by quantifying how faithfully the requested subjects, counts, colors and spatial relations appear, using the same detection and tracking with thresholds calibrated on negative controls. Visible Interaction Error Rate (VIER) assesses physical plausibility through visible interaction errors such as interpenetration, floating and implausible contact. Since such errors resist automatic evaluation, 20 raters judge them blind, and VIER is the percentage of rater--video judgments that report at least one visible interaction error. We further use VBench \cite{vbench} Aesthetic Quality, Subject Consistency and Motion Smoothness to evaluate general video quality. Details of all metrics, together with blind human preference rates for overall quality and task completion, are given in Section A of the appendix.

\begin{figure}[t]
\centering
\includegraphics[width=\columnwidth]{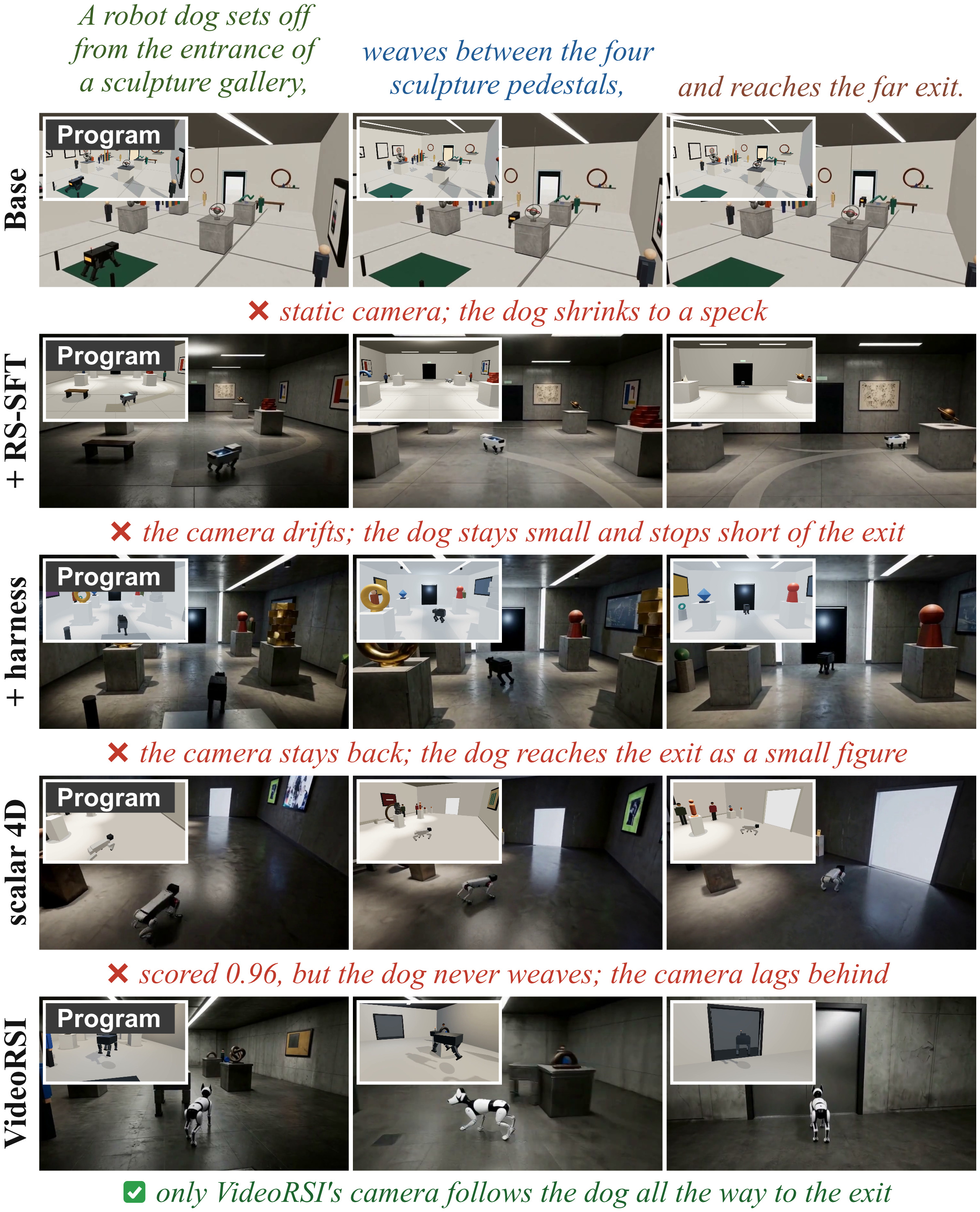}
\caption{Ablation of weights, harness and scalar 4D ranking. Only the full method weaves between the pedestals with a following camera and no scene simplification.}\label{fig:5}
\end{figure}

\noindent\textbf{Quantitative Results.} As shown in Table~\ref{tab:1}, VideoRSI substantially improves CIA and SSF and reduces VIER over Base, while also maintaining competitive perceptual quality. The simultaneous reduction in interaction errors and improvement in scene fidelity is particularly notable: VideoRSI produces fewer spatiotemporal violations while more completely preserving the requested subjects, actions, and spatial relations. This pattern aligns with our selection mechanism: per-dimension 4D constraints limit undesirable trade-offs across quality dimensions, while task-fidelity selection keeps optimization grounded in the complete scene intent. Since all methods share the same frozen video generator, these final-video gains reflect more effective director control over transferable spatiotemporal structure. Improvements from weight learning and harness optimization therefore extend beyond program-level scores to the rendered output. VideoRSI is on par with GPT-6.1 on all three controllability metrics, trailing it only slightly in aesthetic quality. This demonstrates the value of task-specific executable feedback in narrowing the capability gap for a 27B open-source director. Blinded human evaluation supports the same trend; the preference rates are reported in Section A of the appendix.

\noindent\textbf{Qualitative Results.} Fig.~\ref{fig:4} compares how the methods jointly control scene and camera under two multi-stage instructions. In the sculpture gallery, the robot dog must leave the entrance, weave between four pedestals and reach the far exit while the camera follows throughout. Base keeps a static camera as the dog walks straight ahead, failing both weaving and tracking. GPT-6.1 uses a fixed overhead view that hardly shows the route, and its later shots lose the subject, leaving the dog stopped mid-gallery. Video-Coder tilts and collapses pedestals midway and ends facing a blank wall. VideoRSI keeps the pedestals intact and follows the dog to the exit, completing each stage in order and coordinating scene geometry, subject motion and camera trajectory over a complex layout and a long path. In the motorcycle scene, the camera must track from directly behind at wheel height, then close in as the bike brakes. Base stays off to the side; GPT-6.1 holds a high three-quarter view at a fixed distance; Video-Coder tracks from behind but never pushes in. VideoRSI meets the rear position, the wheel height and the push-in during braking, combining subject-relative placement with camera motion that follows the event. These cases show that a director improved with executable 4D feedback keeps scene structure, subject motion and camera behavior coordinated through every stage of the instruction in the final video.

\begin{table*}[t]
\centering
\caption{Final-video controllability and perceptual quality. CIA, VIER and SSF are in \%; the best result in each column is in bold.}\label{tab:1}
\vspace{-4pt}
\small\setlength{\tabcolsep}{0pt}
\renewcommand{\arraystretch}{1.3}
\begin{tabular*}{\textwidth}{@{\extracolsep{\fill}}lcccccc@{}}\toprule
\multirow{2}{*}{Director} & \multicolumn{3}{c}{Controllability} & \multicolumn{3}{c}{Perceptual quality (VBench)} \\
\cmidrule(lr){2-4}\cmidrule(l){5-7}
 & CIA $\uparrow$ & VIER $\downarrow$ & SSF $\uparrow$ & Aesthetic $\uparrow$ & Subject Cons. $\uparrow$ & Motion Smooth. $\uparrow$ \\
\midrule
Base (27B) & 36.8 & 26.8 & 57.6 & 53.9 & 85.6 & 98.1 \\
Video-Coder & 43.8 & 23.6 & 64.9 & 55.6 & 87.8 & 98.6 \\
GPT-6.1 & 51.7 & 18.4 & 72.1 & \textbf{57.9} & 89.4 & 99.0 \\
\oursrow\textbf{VideoRSI} & \textbf{52.6} & \textbf{17.3} & \textbf{73.2} & 56.1 & \textbf{89.7} & \textbf{99.1} \\
\bottomrule\end{tabular*}
\end{table*}

\begin{table}[t]
\centering
\caption{Ablation of VideoRSI components (\%). Best in bold.}\label{tab:3}
\vspace{-4pt}
\small\setlength{\tabcolsep}{0pt}
\renewcommand{\arraystretch}{1.3}
\begin{tabular*}{\columnwidth}{@{\extracolsep{\fill}}lccc@{}}\toprule
Method & CIA $\uparrow$ & VIER $\downarrow$ & SSF $\uparrow$ \\
\midrule
Base (27B) & 36.8 & 26.8 & 57.6 \\
\enspace+ RS-SFT & 42.1 & 25.2 & 60.0 \\
\enspace+ Harness Evolution & 47.8 & 21.7 & 65.9 \\
\midrule
VideoRSI w/ scalar 4D ranking & 48.3 & \textbf{16.8} & 65.5 \\
\oursrow\textbf{VideoRSI (full)} & \textbf{52.6} & 17.3 & \textbf{73.2} \\
\bottomrule\end{tabular*}
\end{table}

\begin{figure}[t]
\centering
\includegraphics[width=\columnwidth]{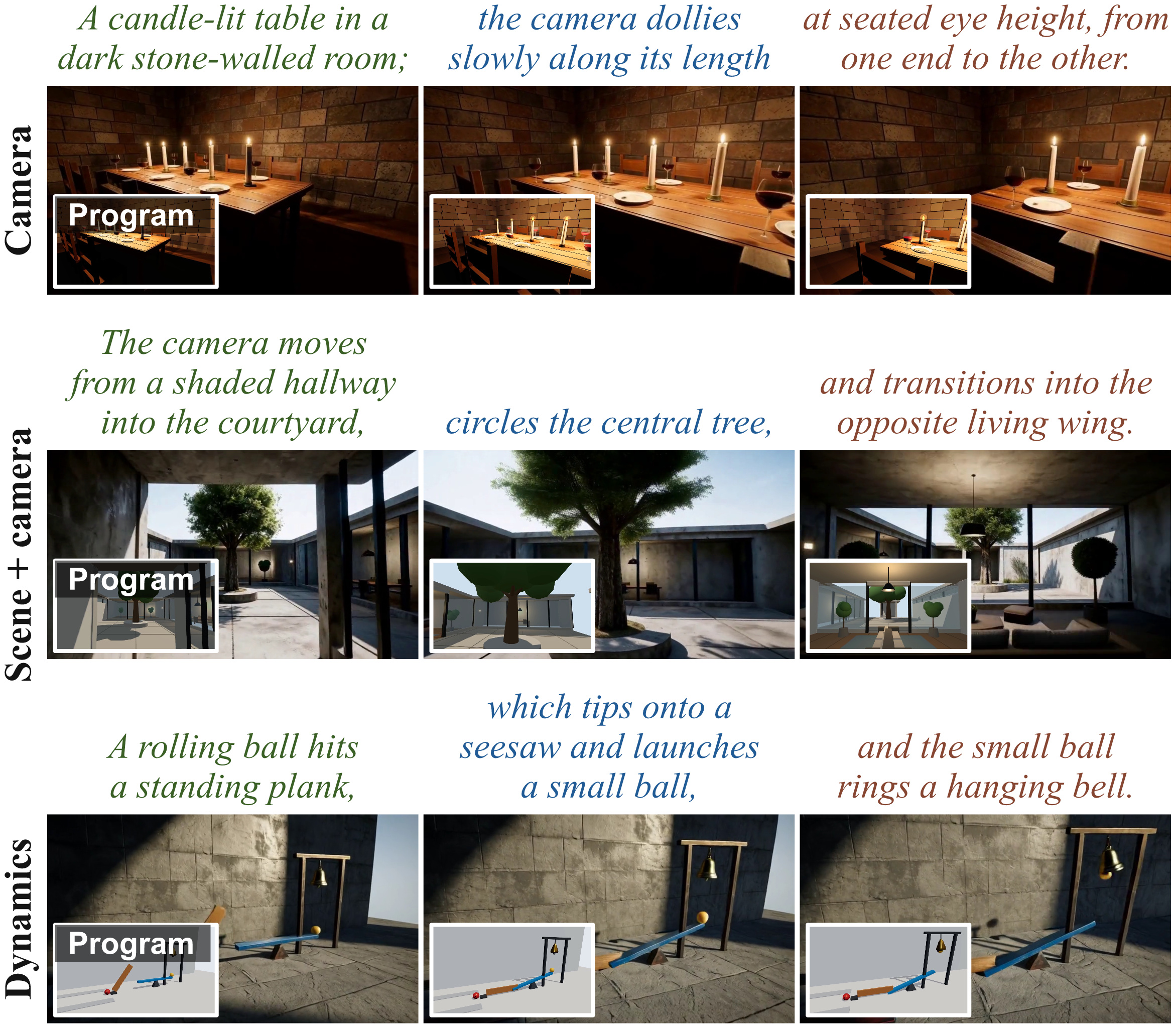}
\caption{General 4D programming: candle-lit camera motion, courtyard traversal and Rube Goldberg interactions.}\label{fig:6}
\end{figure}

\subsection{Ablation Studies}

We ablate RS-SFT, Harness Evolution and constraint-aware selection: each ``+'' row in Table~\ref{tab:3} adds one component to Base, and Fig.~\ref{fig:5} shows qualitative results.

\textbf{Harness Evolution.} The evolved harness organizes executable checks and localized diagnostics into a reusable workflow, so program edits target specific geometric and motion problems: the pedestals stay intact and the dog weaves around them to the exit. Yet the camera stays behind: turning feedback into coordinated camera behavior still depends on the model making effective edits.

\textbf{RS-SFT.} RS-SFT internalizes the construction and repair experience in high-quality trajectories: turn-aligned multimodal supervision links program context, tool feedback and subsequent edits, so the model learns state-dependent execution decisions. With the updated weights the camera starts to follow the subject, yet the shot drifts and the subject stays small and stops short of the exit, so learned experience still needs runtime checking and repair to meet the control requirements of complex scenes.

\textbf{Constraint-aware Selection.} Replacing the full criterion with scalar 4D ranking, all else unchanged, clearly lowers CIA and SSF while VIER stays comparable (Table~\ref{tab:3}). Scalar ranking reaches a slightly lower VIER by favoring programs that delete moving subjects or simplify interactions (Fig.~\ref{fig:5}: a high-scoring program skips the weave); per-dimension constraints with task fidelity account for both plausibility and content completeness.

\textbf{Full VideoRSI.} The full method achieves the best CIA and SSF and follows the dog around the intact pedestals to the exit. Constraint-aware selection keeps correct and faithful experience, RS-SFT learns the construction and repair decisions within it, and the evolved harness supplies executable feedback and workflow support, so training data, model policy and execution workflow jointly coordinate scene geometry, subject motion and camera trajectory.

\subsection{Beyond Fidelity: Direction and Composition}

\looseness=-1 \textbf{Coordinated Scene Direction.} Fig.~\ref{fig:6} shows how VideoRSI jointly directs scene, action and camera: moving along a long table at seated eye height, crossing a courtyard and circling a tree into the opposite room, and organizing continuous multi-object interactions. We orchestrate these elements jointly before generation and use execution feedback to check spatial conflicts and motion readability, so that each camera choice is grounded in the scene. These results extend the focus of controllable generation to the relations among control requirements: camera placement, path and action jointly serve a complete viewing experience, giving the community a concrete path from single controls toward holistic scene direction.

\begin{figure}[t]
\centering
\includegraphics[width=\columnwidth]{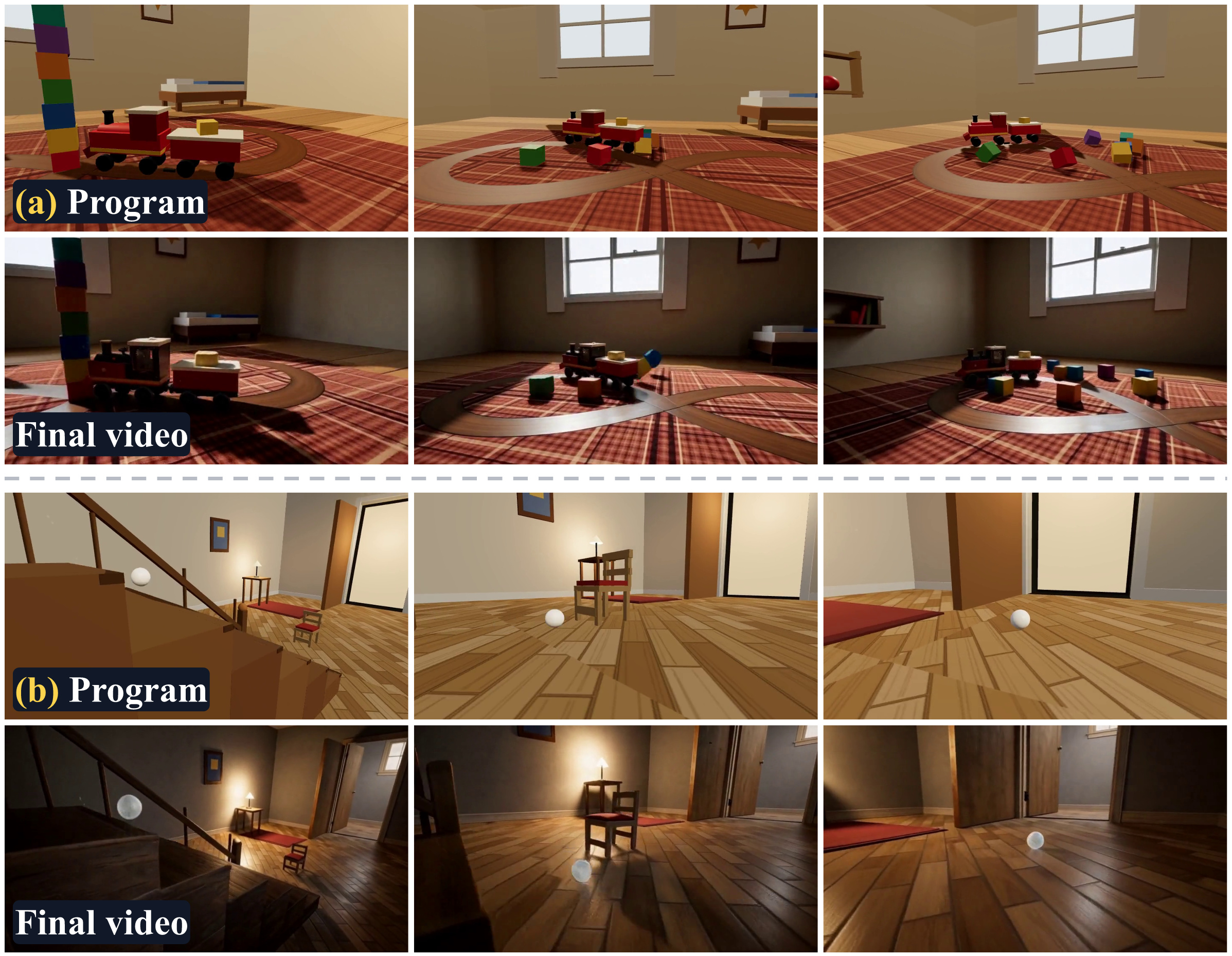}
\caption{Compositional 4D directing: (a) a toy train runs along a figure-eight track and knocks over a block tower as the camera follows; (b) a ball bounces downstairs and hits a chair.}\label{fig:7}
\end{figure}

\looseness=-1 \textbf{Compositional Event Generation.} Fig.~\ref{fig:7} shows composition across space and events. In (a), a toy train runs along a figure-eight track and crashes into a block tower, scattering the blocks across the rug, while a low camera follows the train and pulls back to reveal the aftermath; scene layout, camera motion, track-constrained motion and contact dynamics are realized together. In (b), a ball bounces down the stairs, hits a chair and rolls into the hall. Such processes are linked by dependencies between stages: the outcome of one event is the starting point of the next. RS-SFT learns from successful trajectories to make construction and repair decisions from scene state and execution feedback, which supports this cross-stage organization. The correspondence between program and final video at key events shows how local actions compose into a coherent event chain, providing a compositional basis for more complex visual narratives.

\section{Conclusion}

In this work, we bridge the verification gap in code-to-video generation with VideoRSI, a verifiable self-improvement framework that lets open-source directors learn from their own programs. Our verifier obtains reliable 4D feedback by maintaining a unified 4D state of every object and the camera, gating each spatiotemporal dimension separately and checking the program where its 4D structure is defined, beyond the blind spots of frame-based judges. Driving both RS-SFT and harness evolution with this signal makes self-improvement effective, yielding better videos without simplifying scenes. By verifying what the director controls rather than rendered frames, VideoRSI is a step towards self-improving agents for complex visual generation and brings open-source directors closer to frontier commercial models.

\section*{LLM Usage Statement}

LLMs are part of the studied system: the code-to-video director is an open-weight LLM (Qwen3.8-27B), GPT-6.1 serves as a baseline director, and a multimodal LLM judge is studied as a feedback source. In preparing the paper, GPT-6 \mbox{(OpenAI)} was used only to assist with refining author-written text and code. All methods, experiments, results and claims were designed, run and verified by the authors.

{\small
\bibliographystyle{ieeenat_fullname}
\bibliography{references}
}
\newcommand{\supppage}[1]{\clearpage\thispagestyle{empty}%
  \AddToShipoutPicture*{\put(0,0){\includegraphics[page=#1,width=\paperwidth,height=\paperheight]{supp.pdf}}}%
  \mbox{}}
\supppage{1}
\supppage{2}
\supppage{3}
\supppage{4}
\supppage{5}
\supppage{6}
\supppage{7}
\supppage{8}
\clearpage
\end{document}